\documentclass{article}

\usepackage[utf8]{inputenc} 
\usepackage[T1]{fontenc}    
\usepackage{xcolor}         
\usepackage{microtype}
\usepackage{graphicx, graphics, wrapfig}
\usepackage{subfigure}
\usepackage{amssymb, bm, amsthm}
\usepackage{algorithm, algorithmic}
\usepackage{multirow}
\usepackage{enumitem}
\setlist[itemize,1]{leftmargin=*,labelindent=1.5mm,labelsep=1.mm,itemsep=0.18\baselineskip}
\usepackage{booktabs} 
\usepackage{pifont}
\usepackage{multibib}
\newcites{si}{Additional References for the Supplementary}
\usepackage{amsmath}
\usepackage{accents}
\newlength{\dhatheight}

\usepackage{balance}
\usepackage[font=small,labelfont=bf]{caption}

\newtheorem{innercustomthm}{Theorem}
\newenvironment{customthm}[1]
  {\renewcommand\theinnercustomthm{#1}\innercustomthm}
  {\endinnercustomthm}
\usepackage{hyperref}

\usepackage{mathtools}
\usepackage{nccmath} 

\makeatletter

\newcommand*{\rom}[1]{\expandafter\@slowromancap\romannumeral #1@}
\makeatother

\usepackage[preprint]{neurips_2022}

\newcommand{\modelnameA}{BestowGNN}
\newcommand{\cmark}{\ding{51}}%
\newcommand{\xmark}{\ding{55}}%
\newcommand{\renyi}{R$\acute{\textnormal{e}}$nyi}
\newtheorem{theorem}{Theorem}
\newtheorem{definition}{Definition}
\newtheorem{proposition}{Proposition}

\input{def.set}

\title{A Robust Stacking Framework for Training Deep Graph Models with Multifaceted Node Features}

\author{%
  Jiuhai Chen \thanks{Work done during internship at Amazon Web Services Shanghai AI Lab} \\
  University of Maryland \\
   \And
   Jonas Mueller \& Vassilis N. Ioannidis \\
Amazon \\
   \AND
    Tom Goldstein\\
   University of Maryland  \\
   \And
    David Wipf \\
    Amazon  \\
}

\begin{document}

\maketitle

\begin{abstract}
Graph Neural Networks (GNNs) with numerical node features and graph structure as inputs have demonstrated superior performance on various supervised learning tasks with graph data. However the numerical node features utilized by GNNs are commonly extracted from raw data which is of text or tabular (numeric/categorical) type in most real-world applications. 
The best models for such data types in most standard supervised learning settings with IID (non-graph) data are not simple neural network layers and thus are not easily incorporated into a GNN. Here we propose a robust stacking framework that fuses graph-aware propagation with arbitrary models intended for IID data, which are ensembled and stacked in multiple layers. Our layer-wise framework leverages bagging and stacking strategies to enjoy strong generalization, in a manner which effectively mitigates label leakage and overfitting. Across a variety of graph datasets with tabular/text node features, our method achieves comparable or superior performance relative to both tabular/text and graph neural network models, as well as existing state-of-the-art hybrid strategies that combine the two. 
\end{abstract}

\section{Introduction}
\label{sec:intro}
Graph datasets comprise nodes, which wrap a range of data types and modalities, and edges, which represent the conditional dependence between node feature values. It is commonly assumed that graph neural networks (GNN) are more suitable for such data than models intended for IID data. However GNNs have various limitations, in particular they require appropriately setting many hyperparameters and are nontrivial to apply to graph data in which the node features are not completely numerical. For graph data with tabular (numeric and categorical) features, it has been observed that certain models for IID data (which ignore the graph structure) can be competitive with GNNs, if these models are combined with simple propagation operations to account for the graph structure \citep{huang2020combining, chen2021convergent}. We use the shorthand \emph{IID models} to refer to these models intended for IID data, which in our setting simply operate on one node's features as if they were independent of the other nodes' features (of course this independence does not actually hold for graph data).

Real-world applications of ML typically involve more than just a single model. Instead they require an ML pipeline composed of data preprocessing and  training/tuning/aggregation of many models to achieve the best results. 
In this paper, we investigate how to adapt ML pipelines  designed for supervised learning with IID data to node classification/regression tasks with graph-structured statistical dependence between node features. We focus on using $k$-fold bagging \citep{breiman1996bagging}, i.e.\ \emph{cross-validation},  with stack ensembling \citep{wolpert1992stacked, van2007super}. These techniques are particularly effective for achieving high accuracy across diverse IID datasets, and are utilized in popular AutoML frameworks \citep{erickson2020autogluon}. 

We propose an adaptation of these powerful ensembling techniques for graph data, which aggregates arbitrary IID models that ignore the graph and accounts for the graph structure solely through propagation steps applied at each layer of a stack ensemble.  
Here our goal is to design a single system that can automatically achieve good accuracy across a wide variety of graph datasets without manual dataset-specific adjustment.  
Our proposed graph AutoML system is applied to data in which nodes have text or tabular (numeric/categorical) features, although it  could be applied to image node features as well (without modification, as it can be used with arbitrary models for IID data). Despite not utilizing GNNs at all, our system performs competitively in all of the various prediction tasks we tried.
The contributions of this work include:
\begin{itemize}
  \item We propose a framework of stack ensembling with graph propagation called \textbf{\modelnameA{}} for Bagged, Ensembled, Stacked Training Of Well-balanced GNNs (see Figure \ref{Figure1}) that can \textit{bestow} arbitrary (non-graph) models intended for IID data with the capability of producing highly accurate node predictions in the graph setting.
  
  \item We use analytical tools from differential privacy to understand how our bagging and stacking strategy can effectively mitigate label leakage and  over-fitting with graph-structured data.
  
  \item Without any dataset-specific manual tweaking, our proposed methodology can  match or outperform bespoke dataset-specific models that top competitive leaderboards for popular node classification/regression tasks.
\end{itemize}


\begin{minipage}{\textwidth}
  \begin{minipage}[b]{0.49\textwidth}
    \centerline{\includegraphics[width=1.1\columnwidth, height=12.5cm]{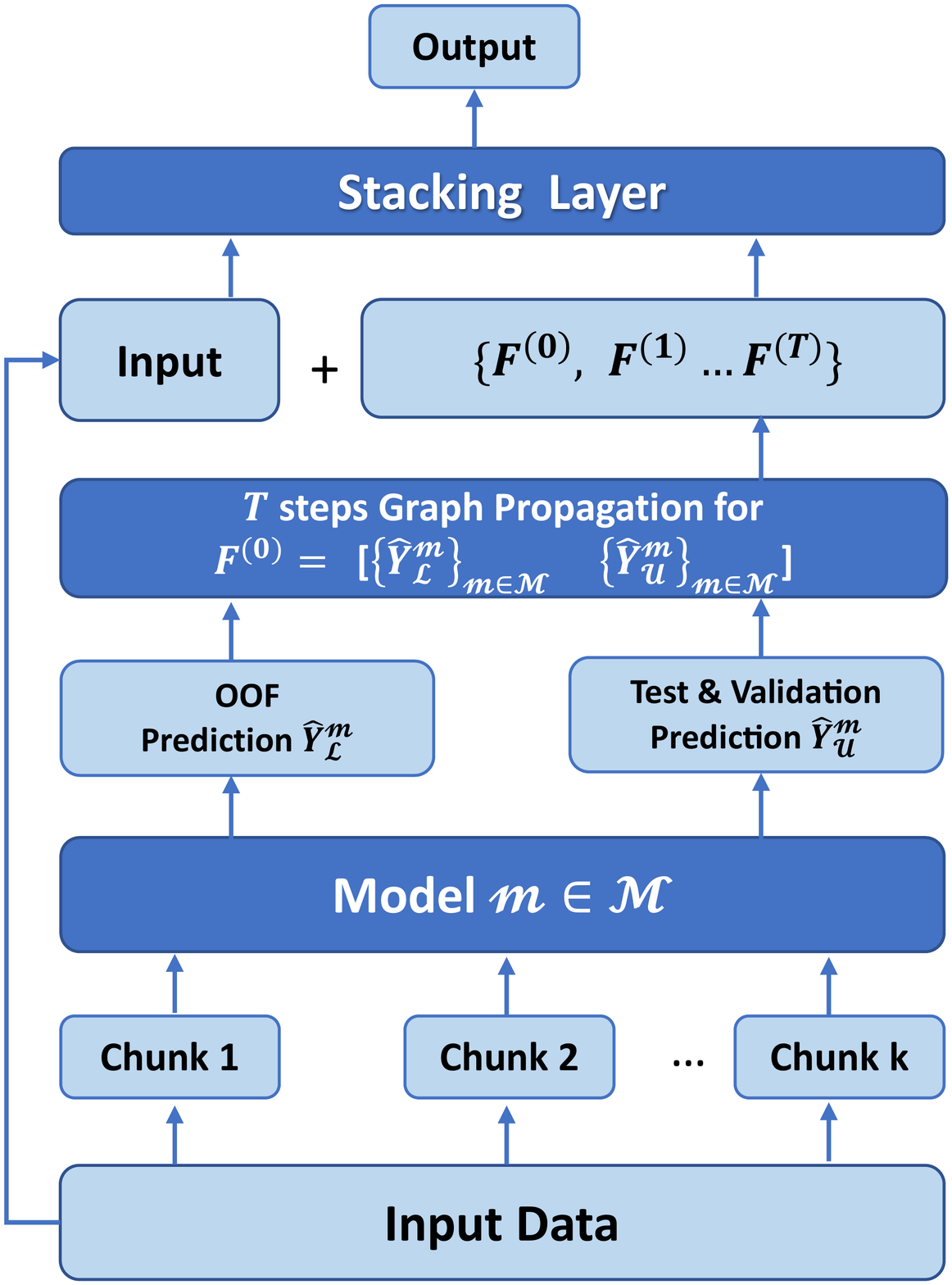}}
    \captionof{figure}{\modelnameA{} with a single base learner, 2 stacking layers, and $k$-fold bagging (repeated bagging not depicted here). The stacking layer repeats the operations depicted between it and the input data.
    }
    \label{Figure1}
  \end{minipage}
  \hfill
  \begin{minipage}[b]{0.49\textwidth}
    \centering
\begin{algorithm}[H]
   \caption{\modelnameA{} Training Strategy}
   \label{alg:bestow}
\begin{algorithmic}
   \STATE {\bfseries Input:} Node features and labels $(\bX, \bY)$ from graph $\mathcal{G}$ with labeled (training) nodes $\calL$ and unlabeled (validation/test) nodes $\calU$, family of models intended for IID data $\mathcal{M}$, $L$ stacking layers, $n$-repeated $k$-fold bagging,  $T$ propagation steps.
   \FOR[stacking]{$l=1$ {\bfseries to} $L$}
   \FOR[repeated bagging]{$i=1$ {\bfseries to} $n$}
   \STATE Randomly split data into $k$ chunks $\{\bX^j, \bY^j\}_{j=1}^k$
   \FOR{$j=1$ {\bfseries to} $k$}
   \STATE   Train model $m\in\mathcal{M}$ on $\{\bX^{-j}, \bY^{-j}\}$
   \STATE Make predictions $\hat{\bY}^j_{m,i}$ on OOF data $\bX^j$
   \ENDFOR
   \ENDFOR
   \FOR{$m \in \mathcal{M}$}
    \STATE Get OOF predictions $\hat{\bY}^m_{\calL}$ for labeled nodes via (\ref{eq:oof_prediction})
    \STATE Get predictions $\hat{\bY}^m_\calU$ for unlabeled nodes via (\ref{eq:test_prediction})
   \ENDFOR
   \STATE Concatenate all models' predictions: \\
   $\bF^{(0)} \triangleq [ \{\hat{\bY}^m_\calL\}_{m \in \calM}, \{\hat{\bY}^m_\calU\}_{m \in \calM} ]$
     \FOR[propagation]{$t=0$ {\bfseries to} $T$}
    \STATE Compute $\bF^{(t)}$ using (\ref{eq:graph-aware-propagation})
    \ENDFOR
    \STATE $\bX\leftarrow$ concatenate $(\bX,\{\bF^{(0)}, ..., \bF^{(T)}\})$  
    \ENDFOR
    \STATE {\bfseries Output:} weighted prediction $\displaystyle \sum_{m \in \calM} \alpha_m \hat{\bY}^m_\calU$ \\
    with $\{\alpha_m\}$ fitted via Ensemble Selection
\end{algorithmic}
\end{algorithm}
    \end{minipage}
  \end{minipage}

\section{Related Work}
\subsection{From Scalability to Layer-wise Training}
Currently, GNN training suffers from high computational cost with the number of layers growing. To improve the scalability of GNNs, graph sampling scheme GraphSAGE \citep{hamilton2017inductive} is adopted by uniformly sampling a fixed number of neighbours for a batch of nodes. Cluster-GCN \citep{chiang2019cluster} uses graph clustering algorithms to sample a block of nodes that form a dense subgraph and runs SGD-based algorithms on these subgraphs. L$^2$-GCN \citep{you2020l2} proposes a layer-wise training framework by disentangling feature aggregation and feature transformation to reduce time and memory complexity. 

SAGN \citep{sun2021scalable} iteratively trains models in several stages by applying graph structure-aware attention mechanisms on node features and also combines the self-training approach with label propagation to further improve performance. GAMLP \citep{zhang2021graph} proposes two attention mechanisms to explore the relation between features with different propagation steps. Both SAGN and GAMLP achieve state-of-the-art performance on two large open graph benchmarks (ogbn-products and ogbn-papers100M), demonstrating the high scalability and efficiency of layer-wise training strategies. However, SAGN and GAMLP suffer from the risk of label leakage: label information is included in the enhanced training set, and can cause performance degredation if the model extracts and relies on these labels. SAGN empirically shows that enough propagation depth can effectively alleviate label leakage, thus they only use label information at one fixed propagation step. Meanwhile, GAMLP passes label information between propagation steps using residual connections. \citet{wang2021bag} further randomly masks nodes during every training epoch to mitigate label leakage issue. 
 
\subsection{Graph models with  Multifaceted Node Features}

Traditional GNN models are mostly studied for graphs with homogeneous sparse node features. Leading GNN models fail to achieve competitive results for heterogeneous features with tabular or text node features \citep{ivanov2021boost,huang2020combining,chen2021convergent}. To remedy this, \citet{ivanov2021boost} jointly train Gradient Boosted Decision Trees (GBDT) and GNN in an end-to-end fashion, demonstrating a significant increase in performance on graph data with tabular node features. 

 \citet{chen2021convergent} removes the need for a GNN altogether, proposing a generalized framework for iterating boosting with parameter-free graph propagation steps that share node/sample information across edges connecting related samples.

Correct and Smooth (C\&S) \citep{huang2020combining} is a simple post-processing step that applies label propagation to further incorporate graph information into the outputs of a learning algorithm. \citet{chen2021convergent} trains 
Gradient Boosted Decision Trees with label propagation incorporated into the objective function, producing competitive results for graph data with tabular node features.

Because common GNNs take numerical node features as inputs, one must establish a way to extract numerical embeddings from raw data such as text and images. For example, the embeddings of ogbn-arxiv data are computed by running the skip-gram model \citep{mikolov2013distributed}. \citet{chien2021node} proposes self-supervised learning to fully utilizing correlations between graph nodes, and extracts the embedding of three open graph benchmark datasets (ogbn-arxiv, ogbn-products and ogbn-papers100M). \citet{chien2021node} demonstrates the superior performance of these new embeddings for the Open Graph Benchmark datasets. \citet{lin2021bertgcn} proposes BertGCN, which combines the Bert model and transductive learning for text classification in an end-to-end fashion and achieves superior performance on a range of text classification tasks. 


\section{Background}
Consider an undirected graph $\calG =(\calV,\calE)$ with $n=|\calV|$ nodes, the node feature matrix is denoted by $\bX\in\mathbb{R}^{n\times d}$ and the label matrix of the nodes is denoted by $\bY\in\mathbb{R}^{n\times c}$ with $d$ and $c$ being the dimension of features and labels. The unweighted adjacency matrix is $\bA\in\mathbb{R}^{n\times n}$. We only access to the labels of a subset of nodes $\{\by_{i}\}_{{i}\in\calL}$, with $\calL \subset\mathcal{V}$.  Given feature values of all nodes $\{\bx_{{i}}\}_{{i}\in \calV}$, label data $\{\by_{{i}}\}_{{i}\in\calL}$, the connectivity of the graph $\mathcal{E}$, the task is to predict the labels of the unlabeled nodes $\{\by_{i}\}_{{i}\in\calU}$, with $\calU=\calV\setminus\calL$. We denote the labeled dataset $\{\bx_i, \by_i\}_{i\in\calL}$ as $D_\calL$, unlabeled dataset $\{\bx_i\}_{i\in\calU}$ as $D_\calU$.

\subsection{Bagging, Ensembling, and Stacking}


For classification/regression with IID (non-graph) data, bagging, ensembling, and stacking represent practical tools that can be combined in various ways to produce more accurate predictions relative to other strategies across diverse tabular and text datasets \citep{shi2021benchmarking,blohm2020leveraging,yoo2020ensemble,fakoor2020fast,bezrukavnikov2021neophyte,smalldata}. For example, in each stacking layer of an ensemble-based architecture, bagging simply trains the same types of base models with out-of-fold predictions from the previous layer models (obtained via bagging) as extra predictive features.  These base models might include various Gradient Boosted Decision Trees \citep{ke2017lightgbm, dorogush2018catboost}, fully-connected neural networks (MLP), K Nearest Neighbors  \citep{erickson2020autogluon} and pretrained Electra Transformer model \citep{clark2020electra}.  For our purposes herein, we adopt the AutoML package AutoGluon~\citep{erickson2020autogluon}, which is capable of exploiting these techniques while serving open-source code that we can readily adapt to include graph propagation.

\subsection{Graph-Aware Propagation Layers}
Recently there has been a surge of interest in connecting GNN layers with an optimization perspective, for example gradient descent and power iterations. Under this scenario, GNN architectures with layers defined can be viewed as the minimization of a principled class of graph-regularized energy functions \citep{klicpera2018predict,ma2020unified,pan2021a,yang2021graph,zhang2020revisiting,zhu2021interpreting}. Hence GNN training can benefit from the inductive bias afforded by energy function minimizers (or close approximations thereof) whose specific form can be controlled by trainable parameters.

Following \citet{zhou2004learning}, the energy function of graph-aware propagation can be given by 
\begin{equation} \label{eq:basic_objective}
\ell_{Y}(\bY) \triangleq  (1-\lambda)\left\|\bY - m\left(\bX ; \btheta  \right) \right\|_{\calF}^2 + \lambda \mbox{tr}\left[\bY^\top \bL \bY  \right],
\end{equation}
where $\lambda\in(0,1)$ is a weight that determines the trade-off between the two terms. $\bY \in \mathbb{R}^{n\times d}$ is a learnable embedding with $d$-dimensional across $n$ nodes, and $m\left(\bX;\btheta\right)$ denotes a base model (parameterized by $\btheta$) that computes an initial target embedding based on the node features $\bX$. $\bL \in \mathbb{R}^{n\times n}$ is the graph Laplacian of $\calG$, meaning $\bL = \bD-\bA$, where $\bD$ represents the degree matrix. 

Intuitively, the first term of (\ref{eq:basic_objective}) encourages $\bY$ to be close to initial target embedding, while the second term introduces the smoothness over the whole graph. 
On the positive side, the closed-form optimal solution of energy function (\ref{eq:basic_objective}) can be easily derived:
\begin{equation} \label{eq:closed form}
\widetilde{m}^*\left( \bX; \btheta \right) \triangleq \arg \min_{\bY} \ell_{Y}(\bY) = \bP^* m\left(\bX ; \btheta  \right),
\end{equation}
with $\bP^* \triangleq \left(\bI + \lambda \bL \right)^{-1}$.
However, for large graphs the requisite inverse is impractical to compute, and alternatively iterative approximations are more practically-feasible. To this end, we may initialize  as $\bY^{(0)} = m\left(\bX ; \btheta  \right)$, and it follows that $\bZ$ can be approximated by iterative descent in the direction of the negative gradient.  Given that
\begin{equation} 
\frac{\partial \ell_{Y}(\bY) }{\partial \bY} = 2\lambda \bL \bY + 2\bY - 2 m\left(\bX ; \btheta  \right),
\end{equation}
the $k$-th iteration of gradient descent becomes
\begin{equation} \label{eq:basic_grad_step}
\bY^{(k)} = \bY^{(k-1)} - \alpha\left[ \left( \lambda \bL  + \bI\right) \bY^{(k-1)} - m\left(\bX ; \btheta  \right) \right], 
\end{equation}
where $\frac{\alpha}{2}$ serves as the effective step size. Consider that $\bL$ is generally sparse, some modifications such as Jacobi preconditioning may be introduced to speed convergence \citep{axelsson1996iterative,yang2021graph} when compute (\ref{eq:basic_grad_step}).

Furthermore, based on well-known properties of gradient descent, if $k$ is sufficiently large and $\alpha$ is small enough, then
\begin{equation} \label{eq:smoothed_pred_function}
    \widetilde{m}^*\left( \bX; \btheta \right)  ~ \approx ~ \widetilde{m}^{(k)}\left( \bX; \btheta \right) ~ \triangleq ~ \bP^{(k)} \left[  m\left(\bX ; \btheta  \right) \right],
\end{equation}
where the operator $\bP^{(k)}\left( \cdot \right)$ computes $k$ gradient steps via (\ref{eq:basic_grad_step}).  The structure of these propagation steps, as well as related variants based on normalized modifications of gradient descent, equate to principled GNN layers, such as those used by GCN \citep{kipf2016semi}, APPNP \citep{klicpera2018predict}, and many others, which can be trained within a broader bilevel optimization framework as described next.

\section{Stack Ensembling for Graph Data (\modelnameA{})}

For node prediction tasks (either regression or classification), each (non-graph) base model is trained within our \modelnameA{} framework by simply  treating each node and its label as a separate IID training example and fitting the model in the usual manner.  
Such a model may informatively encode tabular or text features from the nodes, but its predictions will be uniformed by the additional information available in the  graph structure. To enhance such models with graph information we utilize  graph-aware propagation.

\subsection{Graph-Aware Propagation}
Let $\hat{\bY}_{\calL}, \hat{\bY}_\calU$ denote the predictions of labeled (i.e.\ training) nodes and unlabeled (i.e.\ validation/test) nodes, respectively. In node classification tasks, these may be predicted class probability vectors. 
Via iterative application of the update in (\ref{eq:basic_grad_step}), we can apply graph-aware propagation to predictions $\{\hat{\bY}_{\calL}, \hat{\bY}_\calU\}$ in order to ensure they reflect statistical dependencies between nodes encoded by the graph structure. 
We denote $\bF^{(0)} \triangleq \{\hat{\bY}_\calL, \hat{\bY}_\calU\}$, and for each propagation step $t$: 
\begin{equation}\label{eq:graph-aware-propagation}
    \bF^{(t)} = (1-\lambda) \bF^{(0)} + \lambda\bL \bF^{(t-1)} 
\end{equation}
contains graph-smoothed predictions for the training and test nodes. In our method, $\hat{\bY}$ may actually be predictions from multiple models concatenated together at each node, but the propagation procedure remains identical in this case.

\subsection{Stack Ensembling}
In stack ensembling, the predictions output by individually trained \emph{base}  models are concatenated together as features that are subsequently used to train a \emph{stacker} model whose target is still to predict the original labels    \citep{wolpert1992stacked,ting1997stacking}. A good stacker model learns how to nonlinearly combine the predictions of base models into an even more accurate  prediction. This process can be iterated in multiple layers, a strategy that has been used to win high-profile prediction competitions with IID data \citep{netflix}. 

In this work, we closely follow the stacking methodology of  \citet{erickson2020autogluon}, but adapt it for graphs rather than IID data. We allow stacker models to access the original node features $X$ by concatenating $X$ with the base models' predictions when forming the features used to train each stacker model.
To produce a final prediction for each node, we aggregate predictions from the topmost layer models via a simple weighted combination where weights are learned  via the efficient Ensemble Selection technique of \citet{caruana2004ensemble}. 
Our base models before the first stacking layer are those which can effectively encode the original tabular or text features observed at the nodes (here we utilize AutoGluon which leverages models like Gradient Boosted Decision Trees for tabular features and Transformers for text features). Our stacker models are simply chosen as the same types of models as the base models.

\subsection{Repeated k-fold Bagging to Mitigate Over-fitting}

A problem that arises in the aforementioned stacking strategy is \emph{label leakage}. If a base model is even slightly overfit to its training data such that its predictions memorize parts of the training labels, then subsequent stacker models will have low accuracy due to distribution shift in their features between training and inference time (their features will be highly correlated with the labels during training but not necessarily during inference). This issue is remedied by ensuring stacker models are only trained on features comprised of base model predictions on held-out nodes omitted from the base model's training set. 

We achieve this while still being able to train stacker models using all labeled nodes by leveraging $k$-fold bagging (i.e.\ cross-validation) of all models \citep{van2007super,parmanto1996reducing,erickson2020autogluon}. 
Here the training nodes are partitioned into $k$ disjoint chunks and $k$ copies of each (non-graph-aware) model $m$ are trained with a different data chunk held-out $\{\bX^{-j}, \bY^{-j}\}_{j=1}^k$ held out from each copy. 
After training all $k$ copies of model $m$, we can produce out-of-fold (OOF) predictions $\hat{\bY}^j_{m}$ for each chunk $\bX^j$ by feeding it into the model copy from which it was previously held-out. Following \citet{erickson2020autogluon}, we repeat this $k$-fold bagging procedure over $n$ different random partitions of the training data to further reduce variance and distribution shift that arises in stack ensembling with bagging. Thus for a labeled training node, the OOF prediction from a model of type $m$ is averaged over $n$ different partitions (this node is held-out from exactly one model copy in each partition): 
\begin{equation}\label{eq:oof_prediction}
\hat{\bY}_\calL=\left\{\frac{1}{n}\sum_{i=1}^n \hat{\bY}_{m,i}^j\right\}_{j=1}^k.
\end{equation}
Since unlabeled (validation/test) nodes were technically held-out from every model copy, we can feed them through any copy without harming stacking performance. For a particular type of model $m$, we simply make predictions $\hat{\bY}_\calU$ for unlabeled nodes by averaging over all $n$ bagging repeats and all $k$ copies of the model within each repeat: 
\begin{equation}\label{eq:test_prediction}
\hat{\bY}_\calU =\frac{1}{kn}\sum_{j=1}^k \sum_{i=1}^n\hat{\bY}_{m,i}^j.
\end{equation}
For IID data, this stack ensembling procedure with bagging can produce powerful predictors, both in theory \citep{van2007super} and in practice \citep{erickson2020autogluon}.

\subsection{Stacking with Graph-Aware Propagation}

To extend this methodology to graph data, our proposed training strategy is precisely detailed in Algorithm \ref{alg:bestow}. The main idea is to apply graph-aware propagation on the predictions of models at each intermediate layer of the stack. Different amounts of propagation lead to different characteristics of the data being captured in the resulting prediction (few steps of propagation means predictions are only influenced by local neighbors, whereas many propagation steps allow predictions to be influenced by more distant nodes as well).  Thus we can further enrich the feature set of our stacker models by concatenating together the predictions produced after different numbers of propagation steps. With this expanded feature set, our stacker models learn to aggregate not only the predictions of different models, but differently smoothed versions of these predictions as well. This allows the stacker model to adaptively decide how to best account for  dependencies induced by the graph structure. 

More precisely, if we let $\bF^{(t)}$ denote the predictions (concatenated across all base model types) for labeled and unlabeled nodes after $t$ smoothing steps, then the feature input to each stacker model is given by the original node features $X$ concatenated with  $[\bF^{(0)}, ..., \bF^{(T)}]$. Here the predictions for labeled nodes are always OOF, obtained via bagging. Another fundamental difference between our approach and stack ensembling in the IID setting is the use of unlabeled (test) nodes at each intermediate layer of the stack. By including unlabeled nodes in the propagation, these nodes influence the features used to train subsequent stacker models at labeled nodes. This can even further reduce potential distribution shift in the stacker models' features between the labeled and unlabeled nodes, which ensures better generalization.


Graph machine learning models for non-IID data typically do not use bagging, seemingly because there has not been a rigorous study on the effect of bagging in relation to propagation models. Furthermore, bagging traditionally serves as a means of variance reduction which only brings limited performance benefits for large datasets \citep{breiman1996bagging}.  In contrast, our stacking framework adopts bagging primarily as a means to mitigate the catastrophic effects of label leakage. While bagging can effectively mitigate label information from being directly encoded in stacker model features in the IID setting, it is not clear whether this property still holds with graph-structured dependence between nodes. A particular concern is the fact that the propagation of base model predictions across the graph implies label information is shared across the $k$-fold chunks used to hold-out some nodes from some models. In the next section, we theoretically study this issue and prove that bagging can still mitigate the effects of label leakage even in the non-IID graph setting. 
Our subsequent experiments (see Table \ref{tab:ablation}) reveal that bagging produces substantial performance gains in practical applications of stack ensembling with graph propagation.


\section{Theoretical Analysis}\label{sec:th}
Label utilization is a common technique in which the outputs of a model are concatenated with input features and then used to train a stacking layer. Unfortunately, layer-wise training with label utilization is susceptible to the label leakage problem. Although prior work \citep{sun2021scalable, zhang2021graph} has mentioned heuristic ways to address label leakage via graph propagation, it is unclear how generally applicable this strategy is in practice.  Moreover, there is a natural trade-off between avoiding label leakage via graph propagation, and well-known oversmoothing effects in GNN models.


In this section we employ a powerful theoretical tool, Differential Privacy \citep{mironov2017renyi}, to showcase the advantage of bagging in our proposed \modelnameA{}. Our analysis will show that \modelnameA{} enjoys strong generalization under the \renyi{} Differential Privacy framework. In fact this is the first work that establishes that bagging in graph predictors is useful and mitigates label leakage. Specifically, \modelnameA{} can preserve the privacy (or information sharing) of labels between bags, that would otherwise be compromised by graph propagation.

To this end, we first introduce the definition of \renyi{} Differential Privacy, which is a relaxation of Differential Privacy based on the \renyi{} Divergence. 
\begin{definition}
(\renyi{} Differential Privacy \citep{mironov2017renyi}). Consider a randomized algorithm $\mathcal{M}$ mapping from $\mathcal{D}$ to real-value $\mathcal{R}$. Such an algorithm is said to have $\epsilon$-\renyi{} Differential Privacy if any $D, D^\prime \in \mathcal{D}$ with $d_H(D, D^\prime)=1$, where $d_H$ is the Hamming distance ($D, D^\prime$ are also referred to as adjacent datasets):
\begin{equation}\label{eq:renyi}
    D_\alpha(\mathcal{M}(D)||\mathcal{M}(D^\prime))
    =\frac{1}{\alpha-1}\log E_{x\sim\mathcal{M}(D^\prime)}\left(\frac{\mathcal{M}(D)}{\mathcal{M}(D^\prime)}\right)^\alpha\leq\epsilon.
\end{equation}
\end{definition}
In plain words, the theory establishes that the output of an algorithm does not change significantly when the data changes slightly. The idea behind this framework is that if each individual data sample has only a small effect on the resulting model, the model cannot be used to infer information about any single individual.

We have the following theorem:
\begin{theorem}
\label{thm:1}
Assume base model $m$ to be a multi-layer (two-layer) perceptron and node features $\bX$ is sampled from a multivariate Gaussian as in \citep{jia2021unifying}:
\[
\bX \sim \mathcal{N}(\0, \bGamma^{-1}), \qquad \bGamma=\alpha \bI_n+\beta \bL,
\]
where $\bI_n$ is an identity matrix and $\bL$ is the normalized graph Laplacian. Here $\alpha$ controls noise level and $\beta$ controls the smoothness over the whole graph. $\bE(x_0; D_\calL)$ and $\bF(x_0; D_\calL)$ are predictions produced by \modelnameA{} for a data point $x_0$ with and without bagging mode, respectively. 
If $\bE$ has sensitivity 1, i.e., for any two adjacent $D, D^\prime \in D:|\bE(x_0;D) - \bE(x_0;D^\prime
)| \leq 1$, then $\bE$ satisfies $\alpha/2\sigma^2$-\renyi{} Differential Privacy, where $\sigma^2$ depends on graph structure $\mathcal{G}$. Meanwhile, $\bF$ has no privacy guarantee, i.e., the \renyi{} differential privacy loss (\ref{eq:renyi}) is unbounded.  
\end{theorem}
The proof of theorem \ref{thm:1} is deferred to the supplementary. Theorem \ref{thm:1} indicates that bagging with graph propagation can well preserve the privacy of $D_\calL=\{\bx_i, \by_i\}_{i\in\calL}$ between different chunks while  non-bagging  would have a high risk of leaking the information of $D_\calL$. For layer-wise training with label utilization, the output of the model $\bE(x_0; D_\calL)$ is concatenated with input features and then used to train next stacking layer, and bagging can effectively mitigate the label leakage issue since the information of true label is well preserved at the first layer, while no-bagging exposes the true label and leads to over-fitting issue for next stacking layer.



\section{Experiments}\label{sec:experiments}

\paragraph{Setup.}  We study the effectiveness of our approach by comparing its  performance against state-of-the-art baselines in node regression and classification tasks. As node regression tasks with tabular node features, we consider four real-world graph datasets used for benchmarking by \citet{ivanov2021boost}: House, County, VK and Avazu. As node classification tasks, we consider two datasets with raw text features from the OGB leaderboard \citep{hu2020open}: OGB-Arxiv and OGB-Products. More details about the datasets are provided in the supplementary. 

We compare our method against various baselines, starting with purely tabular baseline models or language models where the graph structure is ignored. Our first baseline is \textbf{Autogluon} \citep{erickson2020autogluon}, an AutoML system for IID tabular or text data that is completely unaware of the graph structure (here we simply treat nodes as IID). Next, we evaluate the performance of \textbf{AutoGluon + C\&S}, which performs Correct and Smooth \citep{huang2020combining} as a posthoc  processing step on top of AutoGluon's predictions, in order to at least account for the graph structure during inference. 
For node regression tasks, we also consider some popular GNN models: \textbf{GCN} \citep{kipf2016semi}, \textbf{GAT} \citep{velivckovic2017graph}, and a hybrid strategy: \textbf{BGNN} \citep{ivanov2021boost}, which combines Gradient Boosted Decision Trees (also a model intended for IID data) with GNNs via end-to-end training in a manner that is graph-aware. 

For node classification tasks, we consider OGB-Arxiv and OGB-Products with raw text as node features (as opposed to some pre-computed text embeddings as node features such as the low-dimensional homogeneous embedding provided by OGB). We compare with \textbf{GIANT-XRT + MLP}, \textbf{GIANT-XRT + GRAPHSAGE} and \textbf{GIANT-XRT + GRAPHSAINT}, which extracts numerical embeddings from text features via a transformer trained through self-supervised learning and feed these high quality embeddings to a multi-layer perceptron or sampling based GNN model. For the smaller OGB-Arxiv dataset, we also consider standard GNN models: \textbf{GCN} \citep{kipf2016semi}, \textbf{GAT} \citep{velivckovic2017graph}. Finally, we compare against SOTA model for OGB-Arxiv and OGB-Products with GIANT-XRT embedding and low-dimensional homogeneous embedding from OGB leaderboard. To our knowledge, there is not a consistent method with superior performance across each dataset. So we compare our \emph{single} general framework with \emph{different} SOTA models for each dataset to ensure we are competing against the best in each case; i.e., \textit{there is no single existing model that is SOTA across them all}. We evaluate our method \textbf{\modelnameA{}}, which incorporates the graph information through propagation operations in each stacking layer, and a variant, \textbf{\modelnameA{} + C\&S}, which adds another final layer of label propagation. Similar to the graph-aware propagation used in our  intermediate stacking layers, this addition is equivalent to applying Correct and Smooth \citep{huang2020combining} to the outputs from \modelnameA{}, as a simple post-processing step that is natural to utilize given that \modelnameA{} already utilizes other propagation steps based on the graph-structure.

\paragraph{Results.} In Table \ref{tab:regression} we present the results for the node regression task with tabular node features. The baseline GNN models are challenged by the tabular node features. AutoGluon is an ensemble of various base models (e.g., Gradient Boosted Decision Trees, fully-connected neural networks) intended for IID data without considering graph structure. We observe that \textbf{Autogluon + C\&S} outperforms \textbf{Autogluon}, demonstrating that graph information can greatly boost the performance of models intended for IID data. Incorporates the graph structure at each stacking layer, our \textbf{\modelnameA{}} method performs better than \textbf{BGNN} on all datasets except VK. The simple addition of C\&S as a natural \modelnameA{} post-processing step is able to further improve performance, outperforming all baselines on all datasets. 

Table \ref{tab:classification} show the results for node classification with either raw text features or numerical text embeddings provided by OGB. Our method \modelnameA{} outperforms all baselines regardless if they leverage the raw text or  OGB embeddings. Note that OGB-Arxiv and OGB-Products have \emph{different} SOTA models in the OGB leaderboard, for instance: \textbf{AGDN + BoT + self-KD + C\&S} is the best existing model for OGB-Arxiv, \textbf{GAMLP + RLU + SCR + C\&S} is the best existing model for OGB-Products. These SOTA models are  \textbf{\emph{manually tweaked}} to perform particularly well only for one specific dataset. In contrast, \textbf{\modelnameA{} + C\&S} uses essentially the same code to fit all datasets without dataset-specific manual adjustment. Comparison of \textbf{\modelnameA{}} with AutoGluon demonstrates how incorporating graph information at each stacking layer can further improve the node classification performance of this AutoML system. More experiments details and computing cost are deferred to the supplementary.

\paragraph{Ablation.} The key ingredients of our framework are bagging/ensembling and graph propagation, Table \ref{tab:ablation} is a ablation study comparing between bagging and no-bagging mode with different propagation steps. Here we consider OGB-Arxiv and OGB-Products with OGB embedding. Table \ref{tab:ablation} presents bagging modes can outperform no-bagging modes for each propagation step, demonstrating bagging effectively mitigate label leakage and over-fitting issue under graph-aware propagation setting.


\begin{table}[tb!]
\caption{Mean squared error of different methods for four node regression datasets.}
\label{tab:regression}
\vskip 0.15in
\begin{center}
\begin{sc}
\scalebox{0.6}{
\begin{tabular}{lccccr}
\toprule
Data set & House  & County & Vk & Avazu \\
\midrule
GCN   & 0.63 $\pm$ 0.01 & 1.48 $\pm$ 0.08 & 7.25 $\pm$ 0.19 & 0.1141 $\pm$ 0.02\\
GAT   & 0.54 $\pm$ 0.01 & 1.45 $\pm$ 0.06 & 7.22 $\pm$ 0.19 & 0.1134 $\pm$ 0.01\\
\midrule
BGNN  & 0.50 $\pm$ 0.01 & 1.26 $\pm$ 0.08 & 6.95 $\pm$ 0.21 & 0.109 $\pm$ 0.01\\
\midrule
AutoGluon   & 0.618 $\pm$ 0.01 &  1.379 $\pm$ 0.08 &  7.176 $\pm$  0.21 & 0.117 $\pm$ 0.018 \\
AutoGluon + C\&S   & 0.477 $\pm$ 0.01 &  1.162 $\pm$ 0.09 &  6.995 $\pm$  0.21 & 0.107 $\pm$ 0.015 \\
\midrule
\modelnameA{}   & 0.495 $\pm$ 0.009 & 1.270 $\pm$ 0.078 & 7.059 $\pm$ 0.218 & 0.108 $\pm$ 0.016 \\
\modelnameA{} + C\&S & \textbf{0.467 $\pm$ 0.007} & \textbf{1.145 $\pm$ 0.083}  & \textbf{6.918 $\pm$ 0.220} & \textbf{0.105 $\pm$ 0.013}\\
\bottomrule
\end{tabular}}
\end{sc}
\end{center}
\vskip -0.1in
\end{table}

\begin{table}[tb!]
    \caption{Node classification accuracy for OGB-Arxiv and OGB-Products achieved by various methods. Rows labeled TEXT contain methods including SOTA models trained on the raw text features at each node, while those labeled OGB indicate models trained on precomputed text embeddings provided by OGB as node features. \emph{SOTA models vary from each dataset with different embeddings/architectures, but \modelnameA{} has consistently superior performance for each dataset without manual dataset-specific adjustment.}}
    \label{tab:classification}
    \begin{minipage}{.5\linewidth}
      \centering
    \caption*{OGB-Arxiv}
      \label{tab:arxiv}
      \scalebox{0.66}{
        \begin{tabular}{lcc}
        \toprule
        Feature & Method  & Test Acc (Validation) \\
        \midrule
        \multirow{3}{*}{OGB} & GCN & 73.06 $\pm$ 0.24 (74.42 $\pm$ 0.12) \\
        & GAT + C\&S & 73.86 $\pm$ 0.14 (74.84 $\pm$ 0.07) \\
        & AGDN+BoT+self-KD+C\&S & 74.31 $\pm$ 0.14 (75.18 $\pm$ 0.09)  \\
        \midrule
        \multirow{4}{*}{text} & GIANT-XRT+MLP & 73.06 $\pm$ 0.11 (74.32 $\pm$ 0.09) \\
        & GIANT-XRT+graphSAGE & 74.35 $\pm$ 0.14 (75.95 $\pm$ 0.11) \\
        & GIANT-XRT+GCN & 75.28 $\pm$ 0.17 (76.87 $\pm$ 0.04) \\
        & GIANT-XRT+RevGAT+KD & 76.15 $\pm$ 0.10 (77.16 $\pm$ 0.09) \\
        \midrule
        \multirow{2}{*}{text} & AutoGluon  & 73.05 $\pm$ 0.00 (74.33 $\pm$ 0.00) \\
        & AutoGluon + C\&S  & 75.34 $\pm$ 0.00 (76.67 $\pm$ 0.00) \\
        \midrule
        \multirow{2}{*}{text} & \modelnameA{}  & 76.06 $\pm$ 0.03 (77.17 $\pm$ 0.06) \\
        & \modelnameA{} + C\&S & \textbf{76.19 $\pm$ 0.02 (77.25 $\pm$ 0.05)} \\
        \bottomrule
        \end{tabular}}
    \end{minipage}%
    \begin{minipage}{.5\linewidth}
      \centering
        \caption*{OGB-Products}
        \label{tab:products}
        \scalebox{0.66}{
    \begin{tabular}{lcc}
    \toprule
    Feature & Method  & Test Acc (Validation) \\
    \midrule
    \multirow{3}{*}{OGB} & DeeperGCN + FLAG & 81.93 $\pm$ 0.31 (92.21 $\pm$ 0.37) \\
    & GAT + FLAG & 81.76 $\pm$ 0.45 (92.51 $\pm$ 0.06)\\
    & GAMLP+RLU+SCR+C\&S & 85.20 $\pm$ 0.08 (93.04 $\pm$ 0.05)\\
    \midrule
    \multirow{4}{*}{text} & GIANT-XRT+MLP & 80.49 $\pm$ 0.28 (92.10 $\pm$ 0.09) \\
    & GIANT-XRT+graphSAGE & 81.99 $\pm$ 0.45 (93.38 $\pm$ 0.05) \\
    & GIANT-XRT+graphSAINT & 84.15 $\pm$ 0.22 (93.18 $\pm$ 0.04) \\
    & GIANT-XRT+SAGN+SLE & 85.47 $\pm$ 0.29 (-)\\
    \midrule
    \multirow{2}{*}{text} & AutoGluon   & 77.10 $\pm$ 0.06 (91.78 $\pm$ 0.03) \\
    & AutoGluon + C\&S  & 79.03 $\pm$ 0.12 (93.62 $\pm$ 0.03) \\
    \midrule
    \multirow{2}{*}{text} & \modelnameA{}  & 85.37 $\pm$ 0.04 (\textbf{94.18 $\pm$ 0.01}) \\
    & \modelnameA{} + C\&S & \textbf{85.48 $\pm$ 0.03} (93.93 $\pm$ 0.02) \\
    \bottomrule
    \end{tabular}}
    \end{minipage} 
\end{table}

\begin{table}[tb!]
\caption{Ablation study of \modelnameA{} with bagging (\cmark) and without bagging (\xmark). $T$ here counts the number of graph  propagation steps, thus $T=0$ represents a baseline model without any graph propagation that ignores the graph structure.}
\label{tab:ablation}
\vskip 0.15in
\begin{center}
\begin{small}
\begin{sc}
\scalebox{0.83}{
\begin{tabular}{lcc|cc}
\toprule
step $T$ & \multicolumn{2}{c}{Arxiv} & \multicolumn{2}{c}{Products} \\
\midrule
   & \cmark & \xmark & \cmark & \xmark\\
0  & 55.70 $\pm$ 0.33 & 54.14 $\pm$ 0.29 & 62.28 $\pm$ 0.35 & 62.05 $\pm$ 0.19\\
1  & 66.25 $\pm$ 0.27 & 64.57 $\pm$ 0.76 & 74.18 $\pm$ 0.21 & 72.61 $\pm$ 0.66\\
2  & 69.34 $\pm$ 0.16 & 67.37 $\pm$ 0.44 & 77.07 $\pm$ 0.32 & 74.61 $\pm$ 0.58 \\
3  & 70.01 $\pm$ 0.16 & 68.08 $\pm$ 0.74 & 78.11 $\pm$ 0.19 & 75.79 $\pm$ 0.49\\
4  & 70.43 $\pm$ 0.21 & 68.72 $\pm$ 0.63 & 78.76 $\pm$ 0.60 & 76.86 $\pm$ 0.17 \\
\midrule
\bottomrule
\end{tabular}}
\end{sc}
\end{small}
\end{center}
\vskip -0.1in
\end{table}


\section{Discussion}

As some AutoML frameworks for IID data (like AutoGluon) can also handle image data as well as multimodal data jointly containing image, text, and tabular features, the methodology presented in this work remains directly applicable to complex graph data in which the nodes contain features from all three modalities (or some nodes have only text features while others have only image/tabular features). Furthermore, as AutoML frameworks for IID data (like AutoGluon) can be trained on messy data with a single line of code, our proposed methods are also easily applied to raw graph data without preprocessing. Our propagation operations are easily added into frameworks like AutoGluon, and thus our proposed methodology can upgrade AutoML for IID data to AutoML for graph data. 
While this paper specifically adopted the models from AutoGluon, our proposed graph  stacking/bagging framework can utilize arbitrary types of classification/regression models intended for IID data. This allows us to tackle node prediction tasks with a flexible combination of all of the best existing models, regardless whether they are applicable to graph data or not.

\clearpage

{
\small
\bibliographystyle{unsrtnat}
\bibliography{VI}
}

\clearpage
\appendix
\thispagestyle{empty}
\begin{center}
    \textbf{\huge Supplementary Materials}
\end{center}
\vspace{0.2cm}

\section{Proof of Theorem 1.}
\paragraph{Preliminary.} Firstly, we derive the format of $\bE(\bx_0; D_\calL)$ and $\bF(\bx_0; D_\calL)$. Suppose \modelnameA{} randomly splits the labeled nodes $D_\calL$ into 2 disjoint chunks $D_1=\{\bX_1, \bY_1\}, D_2=\{\bX_2, \bY_2\}$. \modelnameA{}  trains a model $m\in\mathcal{M}$ with a different data chunk held-out. Model $m$ is defined by a set of parameters collected in  $\btheta$ namely, which is defined as $m(\bX; \btheta)$.
In the following, we will express the predicted labels from model $m$ under the bagging and non-bagging settings. We compare the predicted labels under both settings and establish that our bagging solution is less amenable to label leakage.

The model $m$ will learn different parameters for each chunk and those are denoted as $\btheta_1$ for the chunk \rom{1} and $\btheta_2$ for the chunk \rom{2}, namely $\btheta_1=\btheta(D_1)$ and $\btheta_2=\btheta(D_2)$. Next, \modelnameA{} produces prediction $\hat{\bY}_1, \hat{\bY}_2$ on out-of-fold data, i.e., $\hat{\bY}_1=m(\bX_1; \btheta_2)$ and $\hat{\bY}_2=m(\bX_2; \btheta_1)$. The prediction for unlabeled nodes is $\hat{\bY}_\calU=\frac{1}{2}[m(\bX_\calU; \btheta_1)+m(\bX_\calU; \btheta_2)]$ as explained in (\ref{eq:test_prediction}). Consider one data point $\bx_0$ from the unlabeled dataset $D_\calU$, the prediction of $\bx_0$ is given by $\hat{\by}_0=\frac{1}{2}[m(\bx_0; \btheta_1)+m(\bx_0; \btheta_2)]$. Next, we perform one step graph-aware propagation on $\hat{\by}_0$. 
\begin{equation}\label{bagging}
\begin{aligned}
     \hat{\by}_0^{(1)} & = \sum_{u\in\mathcal{N}(\bx_0)\cap D_\calU}\hat{\by}_u +
    \sum_{v\in\mathcal{N}(\bx_0)\cap D_1}\hat{\by}_v + \sum_{w\in\mathcal{N}(\bx_0)\cap D_2}\hat{\by}_w \\
    & = \sum_{u\in\mathcal{N}(\bx_0)\cap D_\calU}\frac{1}{2}[m(\bx_u; \btheta_1)+m(\bx_u; \btheta_2)]  + \sum_{v\in\mathcal{N}(\bx_0)\cap D_1}m(\bx_v; \btheta_2) + \sum_{w\in\mathcal{N}(\bx_0)\cap D_2}m(\bx_w; \btheta_1),
\end{aligned}
\end{equation}
where $\hat{\by}_0^{(1)}$ is the aggregated results from one-hop neighbor $\mathcal{N}(\bx_0)$, which may belongs to $D_\calU, D_1$ and $D_2$.

Next, we consider the no-bagging mode, where the predictions of $\bX_1, \bX_2$ are changed into $\widetilde{\bY}_1=m(\bX_1; \btheta_1)$ and $\widetilde{\bY}_2=m(\bX_2; \btheta_2)$. Notice that with bagging mode we use the parameters from a different bag, while without bagging we use the parameters from the same bag. The prediction of the test point ${\bx}_0$ is once again $\widetilde{\by}_0=\frac{1}{2}[m(\bx_0; \btheta_1)+m(\bx_0; \btheta_2)]$, which is identical to the bagging mode. We perform the same graph-aware propagation on $\widetilde{\by}_0$.

\begin{equation}\label{no_bagging}
\begin{aligned}
    \widetilde{\by}_0^{(1)} & = \sum_{u\in\mathcal{N}(\bx_0)\cap D_\calU}\widetilde{\by}_u +
    \sum_{v\in\mathcal{N}(\bx_0)\cap D_1}\widetilde{\by}_v + \sum_{w\in\mathcal{N}(\bx_0)\cap D_2}\widetilde{\by}_w \\
    & = \sum_{u\in\mathcal{N}(\bx_0)\cap D_\calU}\frac{1}{2}[m(\bx_u; \btheta_1)+m(\bx_u; \btheta_2)] 
     + \sum_{v\in\mathcal{N}(\bx_0)\cap D_1}m(\bx_v; \btheta_1) + \sum_{w\in\mathcal{N}(\bx_0)\cap D_2}m(\bx_w; \btheta_2).
\end{aligned}
\end{equation}
{Next, we compare the terms among the predicted labels from the two settings, namely (\ref{bagging}) and (\ref{no_bagging}).}
The first term  $\sum_{u\in\mathcal{N}(\bx_0)\cap D_\calU}\frac{1}{2}[m(\bx_u; \btheta_1)+m(\bx_u; \btheta_2)]$ is the same for (\ref{bagging}) and (\ref{no_bagging}) and  can be cancelled. In order to facilitate the exposition of the theoretical contributions we will define functions for the different terms in (\ref{bagging}) and (\ref{no_bagging}). We define $ \bE(\bx_0; D_{\calL})$, that is a function formulating the relation between training data $D_{\calL}$ and the prediction for test data $\bx_0$ under bagging mode.

\begin{equation}\label{bagging_objective}
    \bE(\bx_0; D_{\calL}) := \sum_{v\in\mathcal{N}(\bx_0)\cap D_1}m(\bx_v; \btheta(D_2))  
     +  \sum_{w\in\mathcal{N}(\bx_0)\cap D_2}m(\bx_w; \btheta(D_1)).
\end{equation}

Similarly, we define the function $ \bF(\bx_0; D_{\calL})$ formulating the relation between training data $D_{\calL}$ and the prediction for test data $\bx_0$ under the no-bagging mode:
\begin{equation}\label{no_bagging_objective}
    \bF(\bx_0; D_{\calL}) : = \sum_{v\in\mathcal{N}(\bx_0)\cap D_1}m(\bx_v; \btheta(D_1))  + \sum_{w\in\mathcal{N}(\bx_0)\cap D_2}m(\bx_w; \btheta(D_2)).
\end{equation}
Notice here $\btheta(D_1)$ is the model parameters of Chunk \rom{1} involving information of true label $\bY_1$. We aim to examine bagging and stacking strategies effectively preserve the information of label $\bY_1$ via introducing randomness to the function $\bE(\bx_0; D_{\calL})$ while $\bF(\bx_0; D_{\calL})$ has high risk of leaking the information of true label $\bY_1$.

To proceed in a quantifiable way, we rely on some preliminary results for \renyi{} Differential privacy and generative model for graph learning algorithms.
\begin{proposition}\label{post-process}
\renyi{} differential privacy is preserved by post-processing \citepsi{mironov2017renyi}. If $F(\cdot)$ has $\epsilon$-\renyi{} Differential Privacy, then for any randomized or deterministic function $g$, $g(F(\cdot))$ satisfies $\epsilon$-\renyi{} Differential Privacy.
\end{proposition}

\begin{proposition}\label{gaussian}
If $f$ has sensitivity 1, i.e., for any pair of adjacent datasets $D, D^\prime \in \mathcal{D}$: $|f(D)-f(D^\prime)|\leq 1$, the Gaussian mechanism $\bG_\sigma f$ is said to add Gaussian noise $\mathcal{N}(0, \sigma^2)$ on $f$, then Gaussian mechanism $\bG_\sigma f$ satisfies $\frac{\alpha}{2\sigma^2}$-\renyi{} Differential Privacy \citepsi{mironov2017renyi}.
\end{proposition}

\begin{proposition}\label{conditional}
Consider a multivariate Gaussian distribution, and the random variables are partitioned into two groups $(\bz_P, \bz_Q)$, the distribution is block matrix format
\[ 
\left(\begin{matrix}\bz_P\\\bz_Q\end{matrix}\right)\sim \mathcal{N}\left(\begin{matrix}\left[\begin{matrix}\bar{\bz}_P \\ \bar{\bz}_Q \end{matrix}\right],&\left[\begin{matrix}\bGamma_{PP}  & \bGamma_{PQ} \\  \bGamma_{QP} &  \bGamma_{QQ}  \end{matrix}\right]^{-1}\end{matrix}\right),
\]
where $\left[\begin{matrix}\bGamma_{PP}  & \bGamma_{PQ} \\  \bGamma_{QP} &  \bGamma_{QQ}  \end{matrix}\right]$ is precision (inverse covariance) matrix. Then the marginal and conditional distribution can be written as
\begin{equation}
    \bz_P \sim \mathcal{N}\left(\bar{\bz}_P, (\bGamma_{PP}-\bGamma_{PQ}\bGamma_{QQ}^{-1}\bGamma_{QP})^{-1}\right),
\end{equation}
\begin{equation}
    \bz_P|\bz_Q=\bz_Q \sim \mathcal{N}\left(\bar{\bz}_P-\bGamma_{}^{-1}\bGamma_{PQ}(\bz_Q-\bar{\bz}_Q)\right).
\end{equation}
\end{proposition}



\begin{proposition}
Let $\mathcal{G}=(V, E)$ be an undirected graph, where $V$ is the set of $n$ nodes and $E$ is the set of edges. The adjacency matrix of $\mathcal{G}$ is $\bW\in\mathcal{R}^{n\times n}$, the diagonal degree matrix is $\bD\in\mathcal{R}^{n\times n}$. The normalize graph Laplacian can be written as $\bN=\bI-\bD^{-1/2}\bW\bD^{-1/2}=\bI-\bS$. We use $\bX\in\mathcal{R}^{n\times p}$ for the feature matrix, where p is the dimension of features. We assume all vertex features $\bX$ are jointly sampled from a multivariate Gaussian distribution \citepsi{jia2021unifying}, namely
\begin{equation}
\bX \sim \mathcal{N}(\0, \bGamma^{-1}), \qquad \bGamma=\alpha \bI_n+\beta \bN,
\end{equation}
where $\bI_n$ is identical matrix, $\bN$ is normalized graph Laplacian. Here $\alpha$ controls noise level and $\beta$ controls the smoothness over the whole graph. 

\end{proposition}
We now proceed to our specific results in the main paper.
\begin{customthm}{1}
Assume base model $m$ to be a multi-layer (two-layer) perceptron and node features $\bX$ is sampled from a multivariate Gaussian as in \citesi{jia2021unifying}:
\[
\bX \sim \mathcal{N}(\0, \bGamma^{-1}), \qquad \bGamma=\alpha \bI_n+\beta \bL,
\]
where $\bI_n$ is an identity matrix and $\bL$ is the normalized graph Laplacian. Here $\alpha$ controls noise level and $\beta$ controls the smoothness over the whole graph. $\bE(\bx_0; D_\calL)$ and $\bF(\bx_0; D_\calL)$ are predictions produced by \modelnameA{} for a data point $\bx_0$ with and without bagging mode, respectively. 
If $\bE$ has sensitivity 1, i.e., for any two adjacent $D, D^\prime \in D:|\bE(\bx_0;D) - \bE(\bx_0;D^\prime
)| \leq 1$, then $\bE$ satisfies $\alpha/2\sigma^2$-\renyi{} Differential Privacy, where $\sigma^2$ depends on graph structure $\mathcal{G}$. Meanwhile, $\bF$ has no privacy guarantee, i.e., the \renyi{} differential privacy loss (\ref{eq:renyi}) is unbounded.  
\end{customthm}

\begin{proof}
Given the definition of function $\bE$ and $\bF$ above
\begin{equation}\label{bagging_appendix}
\begin{aligned}
    \bE(\bx_0; D_{\calL}) & = \sum_{v\in\mathcal{N}(\bx_0)\cap D_1}m(\bx_v; \btheta(D_2)) + \sum_{w\in\mathcal{N}(\bx_0)\cap D_2}m(\bx_w; \btheta(D_1))\\
    & = \sum_{v\in\mathcal{N}(\bx_0)\cap D_1}m(\bx_v\btheta(D_2)) + \sum_{w\in\mathcal{N}(\bx_0)\cap D_2}m(\bx_w\btheta(D_1)).
    \end{aligned}
\end{equation}
The second equal because of the MLPs assumptions. 

Similarly,
\begin{equation}\label{no_bagging_appendix}
\begin{aligned}
     \bF(\bx_0; D_{\calL}) & = \sum_{v\in\mathcal{N}(\bx_0)\cap D_1}m(\bx_v; \btheta(D_1))
     + \sum_{w\in\mathcal{N}(\bx_0)\cap D_2}m(\bx_w; \btheta(D_2)) \\
     & = \sum_{v\in\mathcal{N}(\bx_0)\cap D_1}m(\bx_v\btheta(D_1))
     + \sum_{w\in\mathcal{N}(\bx_0)\cap D_2}m(\bx_w\btheta(D_2)).
     \end{aligned}
\end{equation}

Firstly, we define the adjacent dataset $D, D^\prime$: assume $D=D_1$, one data point $\{\bx^\prime,\by^\prime\}$ is randomly selected from Chunk \rom{1} and then remove $\{\bx^\prime,\by^\prime\}$ from $D_1$, i.e., $D^\prime=C_1\backslash\{\bx^\prime,\by^\prime\}$. Meanwhile, unlabeled set $C_\calU$ and  $D_2$ keep the same. Our goal is to examine if $\bE$ and $\bF$ would leak the information of $\{\bx^\prime,\by^\prime\}$ when $\{\bx^\prime,\by^\prime\}$ is removed from $D_1$.

Denote $\bx_v, \bx_w$ as training data in chunk \rom{1} and chunk \rom{2}. Assume $\left(\begin{matrix}\bx_v\\\bx_w\end{matrix}\right)$ is drawn from a multivariate Gaussian distribution: 
\begin{equation}
\left(\begin{matrix}\bx_v\\\bx_w\end{matrix}\right)\sim \mathcal{N}\left(\begin{matrix}\left[\begin{matrix}\0 \\ \0 \end{matrix}\right],&\left[\begin{matrix}\bGamma_{vv}  & \bGamma_{vw} \\  \bGamma_{wv} &  \bGamma_{ww}  \end{matrix}\right]^{-1}\end{matrix}\right),
\end{equation}
where $\left[\begin{matrix}\bGamma_{vv}  & \bGamma_{vw} \\  \bGamma_{wv} &  \bGamma_{ww}  \end{matrix}\right]=\alpha \bI + \beta \bN$, $\bI$ is identical matrix, $\bN$ is normalized graph Laplacian, $\alpha$ controls noise level and $\beta$ controls the smoothness over the whole graph. 

From Proposition \ref{conditional}, the condition distribution of $\bx_w$ given $\bx_v=\bx_v$ can be written as 
\[\bx_w|\bx_v=\bx_v \sim \mathcal{N}(-\bGamma_{ww}^{-1}\bGamma_{wv}\bx_v, \bGamma_{ww}^{-1}).
\]
Condition on the data $D_1$, the distribution of $D_2$ is a conditional multivariate Gaussian distribution with mean $-\bGamma_{ww}^{-1}\bGamma_{wv}\bx_v$ and variance $\bGamma_{ww}^{-1}$. Furthermore, 
$\bx_w\btheta(D_1)$ is also generated from a conditional multivariate Gaussian distribution 
\[\bx_w\btheta(D_1)|\bx_v=\bx_v \sim \mathcal{N}(-\bGamma_{ww}^{-1}\bGamma_{wv}\bx_v\btheta(D_1), \bGamma_{ww}^{-1}).
\]
Thus, $\bx_w\btheta(D_1)$ introduces a Gaussian noise into (\ref{bagging_appendix}). According to Proposition \ref{post-process} and \ref{gaussian}, $\bE$ satisfies $\frac{\alpha}{2\sigma^2}$-\renyi{} Differential Privacy, where $\sigma^2$ depends on $\Gamma_{ww}^{-1}$ decided by graph structure. 

Meanwhile, (\ref{no_bagging_appendix}) is deterministic, we manually add Gaussian noise $\mathcal{N}(0, \sigma^2)$ on (\ref{no_bagging_appendix}), then $\bF$ satisfies $\frac{\alpha}{2\sigma^2}$-\renyi{} Differential Privacy, let $\sigma \rightarrow 0$, we have $\frac{\alpha}{2\sigma^2}\rightarrow \infty$, which indicating $\bF$ has no privacy guarantee only except we manually add Gaussian noise on $\bF$.
\end{proof}

\section{Experiment Details}
\subsection{Data descriptions}
\textbf{House}: node features are the property of house, edges connect the neighbors, the task is to predict the price of the house. 
\textbf{County}: each node is a county and edges connect two counties sharing a border, the task is to predict the unemployment rate for a county. 
\textbf{VK}: each node is a person and edges connect two people based on the friendships, the task is to predict the age of each person.
\textbf{Avazu}: each node is a device and edges connect two devices if they appear on the same site with the same application, the target is the click-through-rate of a node. For \textbf{House}, \textbf{County}, \textbf{VK} and \textbf{Avazu} datasets, Training/validation/testing are randomly split with 6/2/2 ratio and all experiments results are averaged over 5 trails. 

OGB-Arxiv and OGB-Products are standard datasets from OGB-leaderboards and all training/validation/testing splits follow the standard data splitting from OGB-leaderboards.

\subsection{Base models}
Specifically, we consider LightGBM boosted Tress (GBM) \citepsi{ke2017lightgbm}, CatBoost boosted trees (CAT) \citepsi{dorogush2018catboost}, fully-connected neural networks (NN), Extremely Randomized Trees (RT), Random Forests (RF), K Nearest Neighbors (KNN), Label Propagation (LP) \citepsi{huang2020combining} and Transformer with electra pretrained model (Text) (Training epoch is 12) \citepsi{clark2020electra}. For the first layer, we keep the typical models, for example, Gradient Boosted Decision Trees for Tabular data, Transformer models for text data. For second stacking layer, we use all of models except extremely low-efficient models for large dataset, for example, KNN and Catboost slow down the training procedure for OGB-products dataset. All details about the base models can be found in table \ref{base_model}. The parameters about all models can be referred to AutoGluon \citepsi{erickson2020autogluon}.

\subsection{Parameters for Graph-aware propagation}
We do graph-aware propagation for the prediction to incorporate the graph structure. Table \ref{Hyperparameters} shows two hyperparameters considered in (\ref{eq:graph-aware-propagation}): weight $\lambda$ and number of propagation step $T$. We also present the hyperparameters for Correct and Smooth in Table \ref{Hyperparameters_CS}.

\subsection{Computing cost}
The computing cost depends on the ensemble models we select (e.g., transformer models can take more computing resources relying on the implementation, including more emsemble models leads to more computing cost). So it's hard to consistently measure the training/inference time or memory consumption. But the computing cost is in a competitive range since the integration of the bagging and ensembling parts key to our model can be efficiently implemented, e.g., via open source packages like AutoGluon that we used. In Table \ref{tab:time}, we present the training time of different datasets with basic ensemble models. For instance, the training time for OGB-products with OGB embeddings is around 800s, while for GraphSage it is about 1000s for 100 epochs.

\begin{table}[tb!]
\caption{Base models}
\label{base_model}
\vskip 0.15in
\begin{center}
\begin{small}
\begin{sc}
\scalebox{1.0}{
\begin{tabular}{ccc}
\toprule
Dataset & First layer  & Second layer\\
\midrule
House/County/VK/Avazu & CAT, GBM, NN & KNN, GBM, RF, RT, CAT, NN\\
\midrule
OGB-Arxiv & Text, GBM, NN & GBM, RF, RT, NN\\
\midrule
OGB-Products & Text, LP & GBM, RF, RT, NN\\
\bottomrule
\end{tabular}}
\end{sc}
\end{small}
\end{center}
\vskip -0.1in
\end{table}

\begin{table}[tb!]
\caption{Hyperparameters}
\label{Hyperparameters}
\vskip 0.15in
\begin{center}
\begin{small}
\begin{sc}
\scalebox{1.0}{
\begin{tabular}{ccc}
\toprule
Dataset & $\lambda$  &  Input for stacking layer\\
\midrule
House/County/VK/Avazu & 0.9 & $ (\bX,\{\bF^{(0)}_m, \bF^{(1)}_m, \bF^{(2)}_m, \bF^{(3)}_m,  \bF^{(4)}_m, \bF^{(5)}_m\})$\\
\midrule
OGB-Arxiv & 0.95 & $ (\bX,\{\bF^{(0)}_m, \bF^{(1)}_m, \bF^{(3)}_m, \bF^{(5)}_m,  \bF^{(7)}_m, \bF^{(9)}_m\})$\\
\midrule
OGB-Products & 0.97 & $ (\bX,\{\bF^{(0)}_m, \bF^{(1)}_m, \bF^{(3)}_m, \bF^{(5)}_m,  \bF^{(7)}_m, \bF^{(9)}_m\})$\\
\bottomrule
\end{tabular}}
\end{sc}
\end{small}
\end{center}
\vskip -0.1in
\end{table}

\begin{table}[tb!]
\caption{Hyperparameters for C\&S}
\label{Hyperparameters_CS}
\vskip 0.15in
\begin{center}
\begin{small}
\begin{sc}
\scalebox{1.0}{
\begin{tabular}{cccccc}
\toprule
Dataset & $\lambda_1$  &  kernel type& $\lambda_2$  &  kernel type & num\_propagation\\
\midrule
House/County/Avazu & 0.8 & DA & 0.5 & DA & 5\\
\midrule
VK & 0.8 & DA & - & - & 5\\
\midrule
OGB-Arxiv & 0.9 & DA & 0.1 & AD & 50 \\
\midrule
OGB-Products & 0.3 & DAD & 0.3 & AD  & 50\\
\bottomrule
\end{tabular}}
\end{sc}
\end{small}
\end{center}
\vskip -0.1in
\end{table}

\begin{table}[tb!]
\caption{Training time tested on AWS g4dn.12xlarge machine.}
\label{tab:time}
\vskip 0.15in
\begin{center}
\begin{small}
\begin{sc}
\scalebox{1.0}{
\begin{tabular}{ccc}
\toprule
Dataset & Base Model  & Time(s)\\
\midrule
House & GBM, NN & 52\\
\midrule
County & GBM, NN & 18\\
\midrule
VK & GBM, NN & 119\\
\midrule
Avazu & GBM, NN & 15\\
\midrule
OGB-Arxiv & NN & 199\\
\midrule
OGB-Products & NN & 837 \\
\bottomrule
\end{tabular}}
\end{sc}
\end{small}
\end{center}
\vskip -0.1in
\end{table}



\bibliographysi{VI}
\bibliographystylesi{abbrvnat}

\end{document}